\documentclass[11pt,a4paper]{article}
\usepackage[hyperref]{naaclhlt2019}
\usepackage{times}
\usepackage{latexsym}
\usepackage{amsmath}
\usepackage{amssymb}
\usepackage{graphicx}
\usepackage{multirow}
\usepackage{adjustbox}
\usepackage{url}

\aclfinalcopy % Uncomment this line for the final submission
\DeclareMathOperator{\gmax}{max}
\DeclareMathOperator{\gmin}{min}

\def\aclpaperid{19} %  Enter the acl Paper ID here

\title{Generating Text through Adversarial Training\\
        using Skip-Thought Vectors}

\author{Afroz Ahamad \\
  BITS Pilani Hyderabad Campus \\
  {\tt afroz.sahamad@gmail.com} 
  }

\date{}

\begin{document}
\maketitle
\begin{abstract}
GANs have been shown to perform exceedingly well on tasks pertaining to image generation and style transfer. In the field of language modelling, word embeddings such as GLoVe and word2vec are state-of-the-art methods for applying neural network models on textual data. Attempts have been made to utilize GANs with word embeddings for text generation. This study presents an approach to text generation using Skip-Thought sentence embeddings with GANs based on gradient penalty functions and f-measures. The proposed architecture aims to reproduce writing style in the generated text by 
modelling the way of expression at a sentence level across all the works of an author. 
Extensive experiments were run in different embedding settings on a variety of tasks including conditional text generation and language generation. The model outperforms baseline text generation networks across several automated evaluation metrics like BLEU-n, METEOR and ROUGE. Further, wide applicability and effectiveness in real life tasks are demonstrated through human judgement scores.

\end{abstract}

\section{Introduction}

Inducing a particular style in generated text is a promising development which can lead to producing acceptable responses in dialogue generation, image captioning and artificial chat bot systems. In unsupervised text generation, estimating the distribution of real text from a corpus is a challenging task. Recent approaches using adversarial training have addressed this issue by trying to overcome the exposure bias that models trained for maximum likelihood suffer from.  This work proposes an approach for text generation using a Generative Adversarial Network (GAN) with Skip-Thought vectors (STGAN). GANs \cite{GAN} are a class of neural networks that explicitly train a generator to produce high-quality samples by pitting the generator against an adversarial discriminative model. GANs output differentiable values and the task of discrete text generation is challenging because of the non-differentiable nature of generating discrete symbols. Hence, in the present work, the GANs are trained with sentence embedding vectors as a differentiable input. The sentence embeddings are produced using Skip-Thought \cite{kiros2015skip}, a neural network model for learning fixed length representations of sentences.

People's way of expression and communication intention is more diverse across utterances than the vocabulary. To imitate this, the proposed STGAN architecture models the variability at the utterance level in a corpus rather than at word or character level. The effectiveness of this approach is evaluated on automated corpus-based metrics: BLEU-n \cite{BLEU}, METEOR \cite{meteor} and ROUGE \cite{ROUGE} using different embeddings: Average GloVe \cite{glove}, Vector Extrema GloVe \cite{glove} and Skip-Thought \cite{kiros2015skip}.
We perform an empirical study with human judgements to assess both the quality and the style reproduction in the generated text.

\section{Related Works}

Deep neural network architectures have demonstrated strong results on natural language generation tasks such as dialogue response generation and machine translation. Early techniques for generating text conditioned on some input information were template or rule-based engines \cite{mcyag}, or probabilistic models 
such as n-gram. In the recent past, state-of-the-art results on these tasks have been 
achieved by recurrent \cite{rgan, rnn} and convolutional neural network models trained 
for likelihood maximization. Very recently, attempts have been made to generate text using purely generative adversarial training \cite{WGAN}. 

Unsupervised learning with deep neural networks in the framework of encoder-decoder models has become the state-of-the-art methods for approaching NLP problems \cite{recent}. Recent text generation models have used a wide variety of GANs such as policy-gradient based sequence generation framework \cite{seqgan}. \newcite{maskGAN} have used an actor-critic conditional GAN to fill in missing text conditioned on the surrounding text for natural language generation tasks. GANs have also been used for text style transfer by \newcite{yang} where language models act as the discriminator and by \newcite{msrchen} with the introduction of a new f-measure termed as feature-mover's distance. 

Using adversaries of word and character level embeddings for text generation has been explored by \newcite{raj}. Models trained using generative adversarial networks or variational autoencoders have been shown to learn representations of continuous structures by leveraging deep latent variables such as text embeddings \cite{arae}.  This work explores injecting sentence embeddings produced using the Skip Thought architecture \cite{kiros2015skip} into GANs in different setups.

\section{Skip-Thought Generative Adversarial Network}

 In literature corpora such as fantasy and science fiction novels, the vocabulary does not vary significantly across 
 the authors, but the manner of expression does, which is intuitively best captured at the level of sentences than words. 
 The approach, hence, that this work takes in generating sentences with the writing style of one author is to make the adversarial model approximate the distribution of all sentences (rather than words or characters) in a latent space using skip-thought architecture. Previous attempts on text generation have used the character and word-level embeddings instead with GANs \cite{raj}.
 
This section introduces Skip-Thought Generative Adversarial Network with a background on neural network 
models that it is based on. The Skip-Thought model \cite{kiros2015skip} produces embedding vectors for sentences
present in training corpus. These vectors constitute the real distribution for the discriminator network. The 
generator network produces sentence vectors similar to those from the encoded real distribution. The generated 
vectors are sampled over the course of training and then decoded to produce sentences using a Skip-Thought decoder conditioned on 
the same text corpus.

\begin{figure*}[h]
\begin{center}
\centerline{\includegraphics[width=0.9\textwidth]{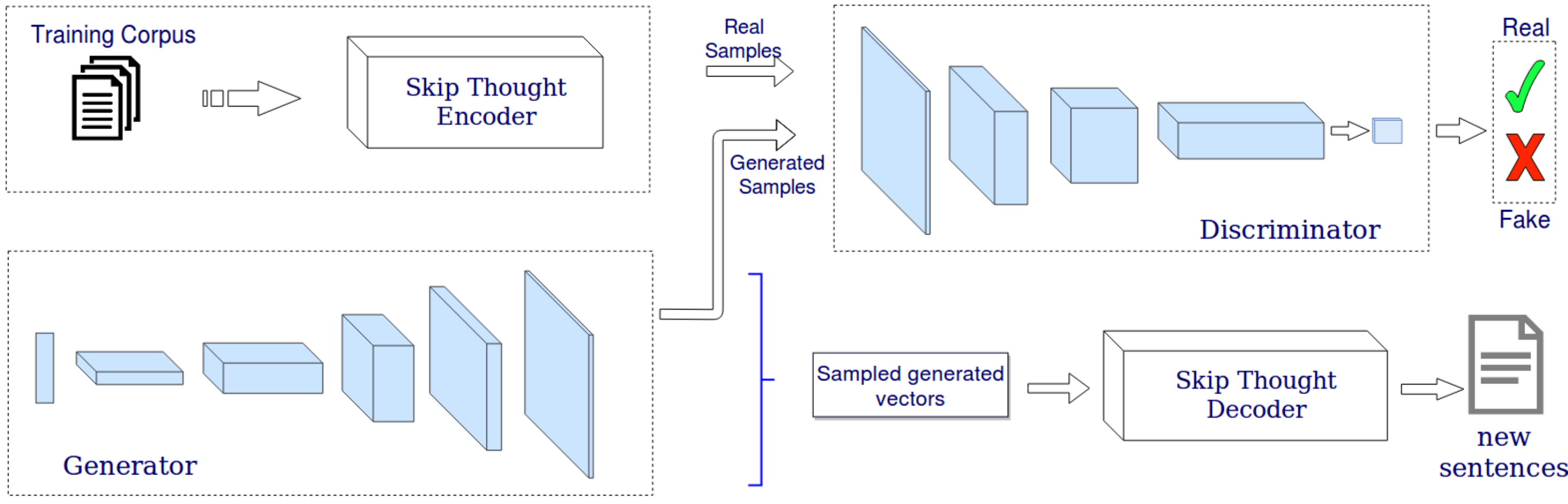}}
\caption{Skip-Thought Generative Adversarial Network model architecture}   
\label{fig:arch}
\end{center}
\vskip -0.2in
\end{figure*} 

\subsection{Skip-Thought Vectors}
Skip-Thought is an encoder-decoder framework with an 
unsupervised approach to train a generic, distributed sentence 
encoder. The encoder maps sentences sharing semantic and syntactic properties to 
similar vector representations and the decoder reconstructs the surrounding sentences 
of an encoded passage. The sentence 
encoding approach draws inspiration from the skip-gram model 
in producing vector representations using previous and next sentences.

The Skip-Thought model uses an RNN encoder with GRU activations \cite{gru}  
and an RNN decoder with conditional GRU. This combination is identical
to the RNN encoder-decoder of \newcite{enc-dec-app} used in neural machine translation.\\

\noindent \textbf{Skip-Thought Architecture}

\noindent For a given sentence tuple $\begin{mathbf}(s_{i-1}, s_{i}, s_{i+1} ) \end{mathbf}$,
let $\begin{mathbf}w_i^t \end{mathbf}$ denote the \textit{t}-th word in sentence $\begin{mathbf}s_{i}\end{mathbf}$ and $\begin{mathbf}x_i^t \end{mathbf}$ denote its word embedding.
The model is described as:

\textbf{Encoder.} Encoded vectors for a sentence \(s_i\) with N words \(w^i\), \(w^{i+1}\),...,\(w^n\) are computed by iterating over the following sequence of equations:
\begin{align*}
\begin{mathbf}{r^{t} = \sigma(W_r x^{t}+U_r h^{t-1})}\end{mathbf}\\
\begin{mathbf}{z^{t} = \sigma(W_z x^{t}+U_z h^{t-1})}\end{mathbf}\\
\begin{mathbf}{\hbar^{t} = \tanh(W x^{t}+U(r^{t} \odot h^{t-1}))}\end{mathbf}\\
\begin{mathbf}{h^{t} = (1-z^{t}) \odot h^{t-1}+ z^{t} \odot \hbar^{t}}\end{mathbf}\
\end{align*}

where \(h_i^t\) is a hidden state at each time step and 
interpreted as a sequence 
of words \(w_i^1\),...,\(w_i^n\), $\begin{mathbf}\hbar^{t}\end{mathbf}$
is the proposed state update at time \textit{t}, $\begin{mathbf}z^{t}\end{mathbf}$
is the update gate and $\begin{mathbf}r^{t}\end{mathbf}$ is the reset gate.
Both update gates take values between zero and one.

\textbf{Decoder.} 
A neural language model conditioned on the encoder output \(h_i\) serves as the decoder. Bias matrices \(C_z\), \(C_r\), \(C\) are introduced for the update gate, reset gate and hidden state computation by the encoder. Two decoders are used in parallel, one each for sentences $\begin{mathbf}s_{i+1}\end{mathbf}$ and $\begin{mathbf}s_{i-1}\end{mathbf}$. 
The following equations are iterated over for decoding:
\begin{align*}
\begin{mathbf}    {r^{t} = \sigma({W_r^d} x^{t-1}+U_r^d h^{t-1} + C_rh_i)}\end{mathbf}\\
\begin{mathbf}    {z^{t} = \sigma(W_z^d x^{t-1}+U_z^d h^{t-1} + C_zh_i)}\end{mathbf}\\
\begin{mathbf}{\hbar^{t} = \tanh(W^d x^{t-1}+U^d(r^{t} \odot h^{t-1}) + Ch_i)}\end{mathbf}\\
\begin{mathbf}{h^{t}_{i+1} = (1-z^{t}) \odot h^{t-1}+ z^{t} \odot \hbar^{t}}\end{mathbf}\
\end{align*}

\textbf{Objective.} 
For the same tuple of sentences, objective function is the sum of log-probabilities 
for the forward and backward sentences conditioned on the encoder representation:\begin{align*}
    \sum_{t} \begin{mathbf} log P( 
    w_{i+1}^t|w_{i+1}^{<t}, h_i)
    \end{mathbf} + \\\sum_{t} \begin{mathbf} log P( 
    w_{i-1}^t|w_{i-1}^{<t}, h_i)
    \end{mathbf}
\end{align*}

\subsection{Generative Adversarial Networks}

Generative Adversarial Networks \cite{GAN}
are deep neural net architectures comprised of two networks,
contesting with each other in a zero-sum game framework.
For a given data, GANs can mimic learning the underlying distribution and generate artificial
data samples similar to those from the real distribution. Generative Adversarial
Networks consists of two players: a Generator and a Discriminator. The generator G
tries to produce data close to the real distribution \(P(x)\) from some  
stochastic distribution \(P(z)\) termed as noise. The discriminator D's
objective is to differentiate between real and generated data \(G(z)\).

The two networks - generator and discriminator compete against 
each other in a zero-sum game. The minimax strategy dictates that each
network plays optimally with the assumption that the other network is optimal. This 
leads to Nash equilibrium which is the point of convergence for GAN model.

\textbf{Objective.}
\newcite{GAN} have formulated the minimax game for a generator G,
discriminator D adversarial network with value function \(V(G,D)\) as:
\begin{multline*}
\underset{G}{\gmin} \underset{D}{\gmax} V(D, G) = {\mathbb{E}}_{x\sim p_{data}(x)}[log D(x)]+\\{\mathbb{E}}_{z\sim p_z(z)}[log(1 - D(G(z)))]
\end{multline*}

\subsection{Model Architecture}

The STGAN architecture (Figure \ref{fig:arch}) has two components: Skip Thought encoder-decoder and a generative adversarial network. The model uses a deep convolutional generative adversarial network, similar to the
one used in DCGAN \cite{dcgan}. During the training, the generator network is updated twice for each discriminator network update to prevent fast convergence of the discriminator network.

The Skip-Thought encoder for the model encodes sentences using 2400 GRU units \cite{gru} with a word vector dimensionality of 620. The encoder combines the sentence embeddings to produce 4800-dimensional combine-skip vectors with the first 2400 dimensions being uni-skip model and the last 2400 bi-skip model. This work uses the 4800-dimensional vectors as they have been found to be the best performing in experiments \footnote{\url{https://github.com/ryankiros/skip-thoughts/}}. For training of the STGAN, the Skip-Thought encoder produces sentence embedding vectors which are labelled as real samples for GAN discriminator.

The decoder uses greedy decoding by taking the argmax over the softmax output distribution for a given time-step which also acts as input for next time-step. It reconstructs sentences conditioned on a sentence vector by randomly sampling from the predicted distributions with a preset beam width. A 620 dimensional RNN word embedding is used for the decoder with 1600 hidden GRU decoding units. All experiments are performed using the Adam optimizer \cite{adam} with gradient clipping and a batch size of 16.

Each sentence is appended with a start token \textless s\textgreater \hspace{0.1cm}  and an end token \textless/s\textgreater  \hspace{0.1cm} before encoding. 
During the process of training generator network with these embeddings, some generated vectors are randomly sampled. The sampled vectors are decoded using pretrained Skip-Thought decoder to produce a probability distribution over the vocabulary in order to reconstruct sentences. The decoding is terminated when the stop token \textless/s\textgreater \hspace{0.1cm} is encountered during reconstruction.

\bgroup
\def\arraystretch{1.7}% 
\begin{table*}[t]
\centering
\begin{sc}
\begin{small}
 \begin{tabular}{|{c}|{c}||{c}|{c}|{c}|{c}|{c}|{c}|} 
 \hline
\textbf{\hfill Model\hfill}  & \textbf{Embedding} & BLEU-1 &  BLEU-2 & BLEU-3 & BLEU-4 & METEOR & ROUGE\\ [0.25ex] 
  \hline
  
\multirow{3}{*}{LSTM} & GloVe Average & 0.874 & 0.792 & 0.621 & 0.582 & 0.681 & \textbf{0.692} \\ \cline{2-8}
                        & GloVe Extreme & 0.874	& 0.791	& 0.616	& 0.580	& 0.677	& 0.685 \\ \cline{2-8}
                        & \textbf{Skip Thought} & \textbf{0.885} & \textbf{0.807}	& \textbf{0.633}	& \textbf{0.585}	& \textbf{0.683}	& \textbf{0.692} \\ \cline{1-8}  
\multirow{3}{*}{Attention BiLSTM} & GloVe Average & \textbf{0.904} & \textbf{0.836}	& 0.645	& 0.583	& \textbf{0.695}	& 0.698 \\ \cline{2-8}
                        & GloVe Extreme & 0.886 & 0.827 & 0.643 & 0.581 & 0.689 & 0.696 \\ \cline{2-8}
                        & \textbf{Skip Thought} & 0.900 &	0.827 & \textbf{0.651} & \textbf{0.589} & 0.692 & \textbf{0.715} \\ \cline{1-8}
\multirow{3}{*}{WGAN -GP} & GloVe Average & 0.879 & 0.807 & 0.668 & 0.585 & \textbf{0.694} & 0.702 \\ \cline{2-8}
                        & GloVe Extreme & 0.853 & 0.799 & 0.666 & 0.579 & 0.689 & 0.697 \\ \cline{2-8}
                        & \textbf{Skip Thought} & \textbf{0.903} & \textbf{0.836} & \textbf{0.682} & \textbf{0.594} & 0.692 & \textbf{0.731} \\ \cline{1-8}
\hline
 \end{tabular}
 \end{small}
 \end{sc}
 \caption{Evaluation of models on word-overlap based automated metrics when trained with different embeddings. Skip-Thought gives better results than GloVe for BLEU-n and ROUGE metrics, while the METEOR scores are comparable to that when using averaged GloVe embedding with Attention BiLSTM generator.}
\label{table:1}
\end{table*}
\egroup

\def\arraystretch{1.4}% 
\begin{table*}[t!]
\centering
\begin{sc}
\begin{small}
 \begin{tabular}{|{c} || {c}|{c} | {c}|{c} | {c}|{c} | {c}|{c}|} 
 \hline
\multirow{2}{*}{\textbf{Model}}  &                 
                  \multicolumn{2}{|c|}{BLEU-2} &
                  \multicolumn{2}{|c|}{BLEU-3} &
                  \multicolumn{2}{|c|}{METEOR} & 
                  \multicolumn{2}{|c|}{ROUGE}\\
                  \cline{2-9}
    & GloVe & \textbf{ST} &  GloVe & \textbf{ST} & GloVe & \textbf{ST} & GloVe & \textbf{ST}  \\
  \hline
 \textbf{GAN}    & 0.710 & 0.745 & 0.593 & 0.607 & 0.667 & 0.670 & 0.654 & 0.649 \\
\textbf{WGAN}    & 0.786 & 0.833 & 0.645 & 0.669 & 0.681 & 0.681 & 0.681 & 0.675 \\
\textbf{WGAN-GP} & 0.807 & 0.836 & 0.668 & 0.682 & 0.694 & 0.692 & 0.702 & 0.731 \\ [1ex] 
\hline
 \end{tabular}
 \end{small}
\end{sc}
 \caption{BLEU-2, BLEU-3 METEOR and ROUGE metric scores across GAN models with different f-measures.
\\
 \textit{\textbf{GloVe}: GLoVe Average, \textbf{ST}: Skip-Thought, \textbf{WGAN}: Wasserstein GAN, \textbf{GP}: Gradient Penalty}}
\label{table:gans}
\end{table*}

\begin{table*}[t!]
\begin{center}
\small
\begin{tabular}{ l|l } 
\hline
\textbf{\hfill Model \hfill} & \textbf{Generated Samples}\\
\hline
{\begin{small}a. \end{small}\textbf{GAN (Mode collapse)}} 
& 1. it ? \\
& 2. it ? \\
& 3. it ? \\
& 4. it ? how would it ? \\
& 5. it ? how would it ? \\ 
\hline
{\begin{small}b. \end{small}\textbf{GAN (minibatch)}} 
& 1. it a bottle ? \\
& 2. a glass bottle ? \\
& 3. a glass bottle it ? \\
& 4. it my hand a bottle ? \\
& 5. the phone my hand it \\
\hline
{\begin{small}c. \end{small}\textbf{Skip Thought WGAN}}
& 1. we have new year ’s holidays, always. \\
& 2. here you can n’t see your suitcase , \\
& 3. please show me how much is a transfer? \\
& 4. i had a police take watch out of my wallet . \\
& 5. here i collect my telephone card and telephone number \\
\hline
{\begin{small}d. \end{small}\textbf{Skip Thought WGAN-GP}}
& 1. my passport and a letter card with my card , please \\
& 2. here on my telephone, mr. kimura’s registration card’s address. \\
& 3. i can n’t see some shopping happened . \\
& 4. get him my camera found a person ’s my watch . \\
& 5. delta airlines flight six zero two from six p.m. to miami, please? \\
\hline
\end{tabular}
\caption{\label{table:2}Sentences sampled from STGAN when training on CMU-SE Dataset;
mode collapse is overcome by using minibatch discrimination. Sample quality in terms of length and diversity further improved by using Wasserstein distance f-measure with gradient penalty regularizer.  \textit{\textbf{WGAN}: Wasserstein GAN,  \textbf{GP}: Gradient Penalty}}
\end{center}
\end{table*}

\subsection{\label{sec:34}Improving Training and Loss}
The training process of a GAN is notably difficult \cite{improveTrainGAN}
and several improvement techniques such as batch normalization, feature matching, historical averaging \cite{improveTrainGAN} and unrolling GAN \cite{unrollGAN} have been suggested for making the training
more stable. 
Training the Skip-Thought GAN often results in mode 
dropping \cite{trainGAN, veeGAN} with a parameter setting where it outputs a very narrow 
distribution of points. To overcome this, it uses minibatch discrimination by 
looking at an entire batch of samples and modeling the distance
between a given sample and all the other samples present in that batch.

The minimax formulation for an optimal discriminator in a vanilla GAN is 
Jensen-Shannon Distance between the generated distribution and 
the real distribution. \newcite{WGAN} used Wasserstein distance or
earth mover's distance to demonstrate how replacing distance measures can improve training loss for a GAN. \newcite{GulWGAN} have incorporated
a gradient penalty regularizer term in WGAN objective for discriminator's 
loss function. The experiments in this work use the above f-measures to improve performance of Skip-Thought GAN on text generation.

\section{Results and Discussion} 
\subsection{Conditional Generation of Sentences.}
GANs can be conditioned on certain attributes to generate real valued data \cite{cgan, dcgan}. In this experiment, both the generator and discriminator are conditioned on the Skip-Thought encoded vectors.

The data used for this setup consists of 250,000 sentences chosen from the BookCorpus dataset
\cite{bookCorpus} with a training/test/validation split of 5/1/1. All the sentences belong to one series of fantasy novels by a particular author of English language. This selection implies that the author's word choice, sentence structure, figurative language, and sentence arrangement are consistent and well-represented across the dataset. Conditioning on this high-level outline gives more robustness to the model in terms of generated samples. 

 The decoded sentences form the evaluation set for measuring performance of different models under corpus level BLEU-n \cite{BLEU}, METEOR \cite{meteor} and ROUGE\cite{ROUGE} metrics. Table \ref{table:1} compares these results for the proposed STGAN against standard LSTM and Attention Bidirectional LSTM models in two settings - using Skip Thought vectors and using tied GloVe \cite{glove} embeddings. For this experiment, GloVe Average is obtained by averaging GloVe embeddings of all the words composing a given sentences while GloVe Extreme is computed by taking the most extreme value for each dimension across embeddings of all the words. All the three models used in the experiment: LSTM, Attention BiLSTM and Wasserstein GAN take the above three embeddings as input in separate runs and output vectors which are decoded to reconstruct sentences using the corresponding embedding's decoder.
 
Table \ref{table:gans} shows improvements in metric scores when using Wasserstein distance and gradient penalty regularizer as discussed in section \ref{sec:34}. WGAN-GP gives the strongest across-the-board performance in both the GloVe and Skip-Thought settings, so we use this as the basis for the rest of our experiments. The METEOR scores are reportedly better for other models because though it does not rely on embeddings but it includes notions of synonymy and paraphrasing to compute alignment between hypothesis and reference sentences \cite{nlgeval}.

\subsection{Language Generation.}
Language generation is performed on a dataset comprising simple English sentences 
referred to as 
CMU-SE\footnote{\url{https://github.com/clab/sp2016.11-731/tree/master/hw4/data}} \cite{raj}. The CMU-SE dataset consists of 44,016 sentences with a vocabulary of 
3,122 words. The vectors are extracted in batches of same-lengthed sentences for encoding. 
The samples represent how mode collapse is manifested when using least-squares distance \cite{lsgan}
f-measure without minibatch discrimination. Table \ref{table:2}(a) contains sentences generated from
a vanilla STGAN which mode collapse is observed, while \ref{table:2}(b) contains examples wherein it is not observed when using minibatch discrimination.
Table \ref{table:2}(c) shows generated samples from STGAN when using Wasserstein distance 
f-measure as WGAN \cite{WGAN}) and \ref{table:2}(d) contains samples when using a gradient penalty regularizer term as WGAN-GP \cite{GulWGAN}. The two models generate longer human-like sentences and over a more diverse vocabulary.

\begin{figure}[b!]
\begin{center}
\centerline{\includegraphics[width=0.5\textwidth]{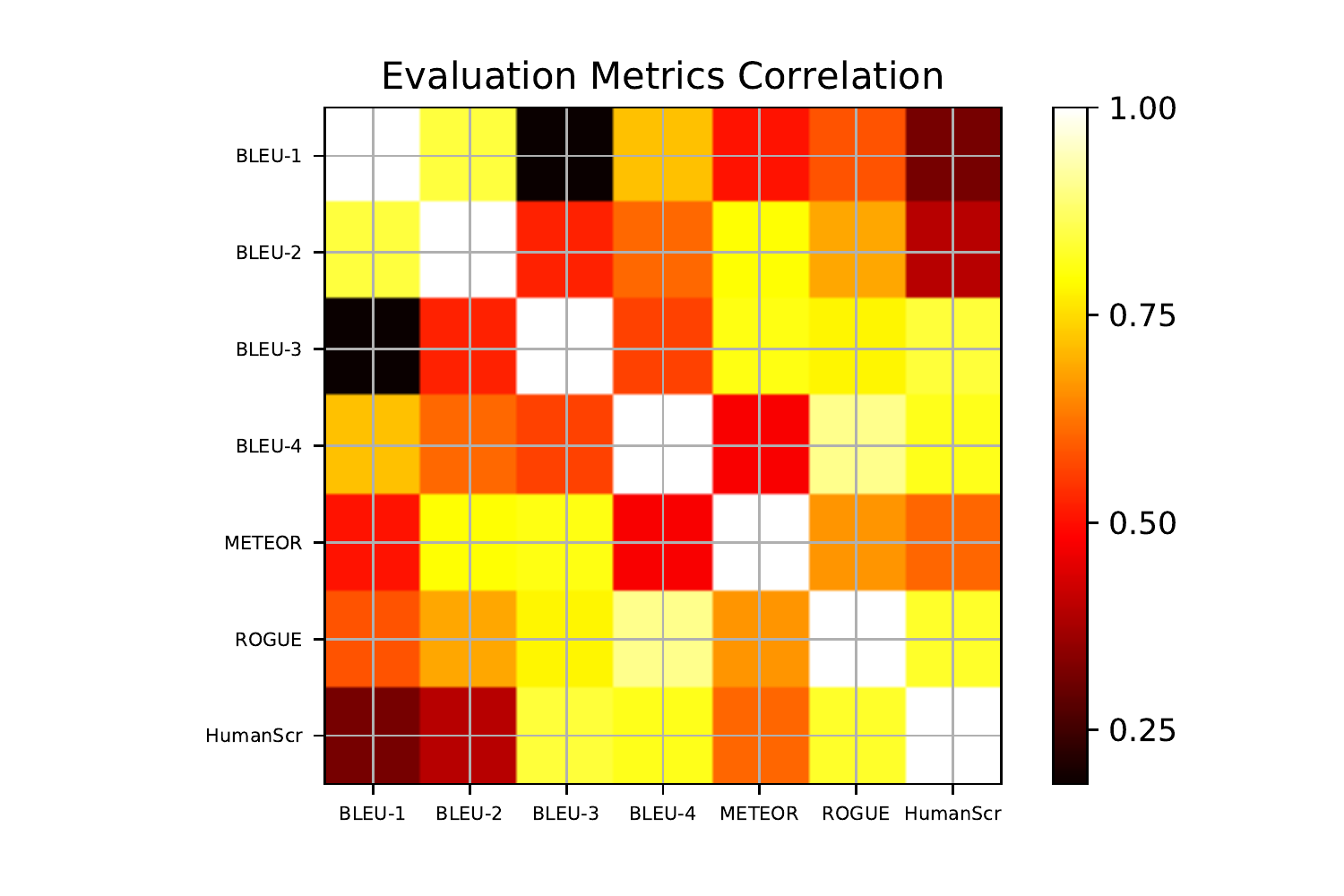}}
\caption{Pearson's correlation coefficient between automated computed metrics and human scores. Human scores correlate well with BLEU-3 and ROUGE scores.}   
\label{fig:pear}
\end{center}
\end{figure}

\begin{table}[h!]
\def\arraystretch{1.4}% 
\centering
\small
 \begin{tabular}{|{c} | {c} | {c} || {c} | {c} |} 
 \hline
  & Real & Fake & \% real & \% fake \\\hline
 Real & 30 & \textbf{51} & 37.04\% & \textbf{62.96}\% \\ \hline
 Fake & \textbf{48} & 75 & \textbf{39.02}\% & 60.98\%\\ \hline
 \end{tabular}
 \caption{Weighted human scores for sentences. $\begin{mathbf} \vert rating-3 \vert \end{mathbf}$ is weight given to each sentence's rating. 39.02\% of the generated samples were marked as real.}
\label{table:human}
\end{table}

\subsection{Human Scores and Correlations}
The performance of this approach to generate new sentences has been
evaluated in reproducing writing style of a particular author. The participant group consisted of 14 individuals, who were 
familiar with writing style of the said author by having read all but a few of the literary works of the author. The 
setup prevents them from being 
certain whether a sentence in question has or has not appeared 
in any work of the author that they have already read. To form the evaluation set of sentences, the generated samples were mixed with real 
sentences from the author's writing. 10 sentences from this mixed pool were chosen at random to be presented to each person. 
The participants were asked to mark on a scale of 1 to 5 if they
thought that a sentence seemed to belong to the 
author's works or was generated from a model, with 1 being 
certainly from the author and 5 being certainly from a model. 
Table \ref{table:human} shows the weighted scores computed as $\begin{mathbf} \vert rating - 3 \vert \end{mathbf}$ to account for the degree of uncertainty addressed by a participant when rating. The models performs well with 39.02\% of the generated samples being marked as written by the author while a greater 62.96\% of the actual sentences from author's writing being marked as fake generated ones. Figure \ref{fig:pear} compiles Pearson's correlation coefficients 
between the obtained human scores and Skip-Thought GAN scores.

\section{Conclusion}
This work presents a simple and effective model for text generation based on adversarial 
training using sentence embeddings. It shows how the use of sentence-level embeddings allows modelling the way of expression of an author in generated text in a better way than when using word-level 
embeddings. A performance comparison across several metrics is made between different GAN 
architectures with improved training stability and attention augmented LSTM models. Finally, it discusses how the automated corpus-based evaluations correlate with human judgements. In 
future, this work aims to be applied for synthesizing images from text, exploring
complementary architectures to projects like 
neural-storyteller\footnote{\url{https://github.com/ryankiros/ne
ural-storyteller}} where skip-thought embeddings are already 
used to perform image captioning with story-style transfer.
\pagebreak

\section*{Acknowledgements}
The author would like to thank Aruna Malapati for providing insights and access to an Nvidia Titan X GPU for the experiments; and Pranesh Bhargava, Greg Durrett and Yash Raj Jain for providing helpful feedback. The author also acknowledges the support of Microsoft Research India Travel Grant.

\bibliography{naaclhlt2019}

\begin{thebibliography}{30}
\expandafter\ifx\csname natexlab\endcsname\relax\def\natexlab#1{#1}\fi

\bibitem[{{Arjovsky} and {Bottou}(2017)}]{trainGAN}
M.~{Arjovsky} and L.~{Bottou}. 2017.
\newblock \href {http://arxiv.org/abs/1701.04862} {{Towards Principled Methods
  for Training Generative Adversarial Networks}}.
\newblock \emph{ArXiv e-prints}.

\bibitem[{{Arjovsky} et~al.(2017){Arjovsky}, {Chintala}, and {Bottou}}]{WGAN}
M.~{Arjovsky}, S.~{Chintala}, and L.~{Bottou}. 2017.
\newblock \href {http://arxiv.org/abs/1701.07875} {{Wasserstein GAN}}.
\newblock \emph{ArXiv e-prints}.

\bibitem[{Banerjee and Lavie(2005)}]{meteor}
Satanjeev Banerjee and Alon Lavie. 2005.
\newblock Meteor: An automatic metric for mt evaluation with improved
  correlation with human judgments.
\newblock In \emph{Proceedings of the ACL Workshop on Intrinsic and Extrinsic
  Evaluation Measures for Machine Translation and/or Summarization}, pages
  65--72. Association for Computational Linguistics.

\bibitem[{Chen et~al.(2018)Chen, Dai, Tao, Shen, Gan, Zhang, Zhang, and
  Carin}]{msrchen}
Liqun Chen, Shuyang Dai, Chenyang Tao, Dinghan Shen, Zhe Gan, Haichao Zhang,
  Yizhe Zhang, and Lawrence Carin. 2018.
\newblock \href {http://arxiv.org/abs/1809.06297} {Adversarial text generation
  via feature-mover's distance}.
\newblock \emph{CoRR}, abs/1809.06297.

\bibitem[{Cho et~al.(2014)Cho, van Merrienboer, Bahdanau, and
  Bengio}]{enc-dec-app}
Kyunghyun Cho, Bart van Merrienboer, Dzmitry Bahdanau, and Yoshua Bengio. 2014.
\newblock On the properties of neural machine translation: Encoder--decoder
  approaches.
\newblock In \emph{Proceedings of SSST-8, Eighth Workshop on Syntax, Semantics
  and Structure in Statistical Translation}. Association for Computational
  Linguistics.

\bibitem[{Chung et~al.(2014)Chung, Gulcehre, Cho, and Bengio}]{gru}
Junyoung Chung, Caglar Gulcehre, KyungHyun Cho, and Yoshua Bengio. 2014.
\newblock Empirical evaluation of gated recurrent neural networks on sequence
  modeling.
\newblock \emph{arXiv preprint arXiv:1412.3555}.

\bibitem[{{Fedus} et~al.(2018){Fedus}, {Goodfellow}, and {Dai}}]{maskGAN}
W.~{Fedus}, I.~{Goodfellow}, and A.~M. {Dai}. 2018.
\newblock \href {http://arxiv.org/abs/1801.07736} {{MaskGAN: Better Text
  Generation via Filling in the\_\_\_\_\_\_}}.
\newblock \emph{ArXiv e-prints}.

\bibitem[{Goodfellow et~al.(2014)Goodfellow, Pouget-Abadie, Mirza, Xu,
  Warde-Farley, Ozair, Courville, and Bengio}]{GAN}
Ian Goodfellow, Jean Pouget-Abadie, Mehdi Mirza, Bing Xu, David Warde-Farley,
  Sherjil Ozair, Aaron Courville, and Yoshua Bengio. 2014.
\newblock Generative adversarial nets.
\newblock In \emph{Advances in Neural Information Processing Systems 27}, pages
  2672--2680. Curran Associates, Inc.

\bibitem[{Gulrajani et~al.(2017)Gulrajani, Ahmed, Arjovsky, Dumoulin, and
  Courville}]{GulWGAN}
Ishaan Gulrajani, Faruk Ahmed, Martin Arjovsky, Vincent Dumoulin, and Aaron~C
  Courville. 2017.
\newblock Improved training of wasserstein gans.
\newblock In \emph{Advances in Neural Information Processing Systems 30}, pages
  5767--5777.

\bibitem[{Kingma and Ba(2014)}]{adam}
Diederik~P. Kingma and Jimmy Ba. 2014.
\newblock \href {http://arxiv.org/abs/1412.6980} {Adam: {A} method for
  stochastic optimization}.
\newblock \emph{CoRR}, abs/1412.6980.

\bibitem[{Kiros et~al.(2015)Kiros, Zhu, Salakhutdinov, Zemel, Torralba,
  Urtasun, and Fidler}]{kiros2015skip}
Ryan Kiros, Yukun Zhu, Ruslan Salakhutdinov, Richard~S Zemel, Antonio Torralba,
  Raquel Urtasun, and Sanja Fidler. 2015.
\newblock Skip-thought vectors.
\newblock \emph{arXiv preprint arXiv:1506.06726}.

\bibitem[{Lin(2004)}]{ROUGE}
Chin-Yew Lin. 2004.
\newblock Rouge: A package for automatic evaluation of summaries.
\newblock \emph{Text Summarization Branches Out}.

\bibitem[{Mao et~al.(2016)Mao, Li, Xie, Lau, and Wang}]{lsgan}
Xudong Mao, Qing Li, Haoran Xie, Raymond Y.~K. Lau, and Zhen Wang. 2016.
\newblock Multi-class generative adversarial networks with the {L2} loss
  function.
\newblock \emph{CoRR}.

\bibitem[{McRoy et~al.(2000)McRoy, Channarukul, and Ali}]{mcyag}
Susan~W McRoy, Songsak Channarukul, and Syed~S Ali. 2000.
\newblock Yag: A template-based generator for real-time systems.
\newblock In \emph{Proceedings of the first international conference on Natural
  language generation-Volume 14}, pages 264--267. Association for Computational
  Linguistics.

\bibitem[{Metz et~al.(2016)Metz, Poole, Pfau, and Sohl{-}Dickstein}]{unrollGAN}
Luke Metz, Ben Poole, David Pfau, and Jascha Sohl{-}Dickstein. 2016.
\newblock Unrolled generative adversarial networks.
\newblock \emph{CoRR}, abs/1611.02163.

\bibitem[{Mikolov et~al.(2010)Mikolov, Karafi{\'a}t, Burget, Cernock{\'y}, and
  Khudanpur}]{rnn}
Tomas Mikolov, Martin Karafi{\'a}t, Luk{\'a}s Burget, Jan Cernock{\'y}, and
  Sanjeev Khudanpur. 2010.
\newblock Recurrent neural network based language model.
\newblock In \emph{INTERSPEECH}.

\bibitem[{Mirza and Osindero(2014)}]{cgan}
Mehdi Mirza and Simon Osindero. 2014.
\newblock Conditional generative adversarial nets.
\newblock \emph{CoRR}, abs/1411.1784.

\bibitem[{Papineni et~al.(2002)Papineni, Roukos, Ward, and Zhu}]{BLEU}
Kishore Papineni, Salim Roukos, Todd Ward, and Wei-Jing Zhu. 2002.
\newblock Bleu: A method for automatic evaluation of machine translation.
\newblock In \emph{Proceedings of the 40th Annual Meeting on Association for
  Computational Linguistics}, ACL '02.

\bibitem[{Pennington et~al.(2014)Pennington, Socher, and Manning}]{glove}
Jeffrey Pennington, Richard Socher, and Christopher Manning. 2014.
\newblock Glove: Global vectors for word representation.
\newblock In \emph{Proceedings of the 2014 conference on empirical methods in
  natural language processing (EMNLP)}, pages 1532--1543.

\bibitem[{Press et~al.(2017)Press, Bar, Bogin, Berant, and Wolf}]{rgan}
Ofir Press, Amir Bar, Ben Bogin, Jonathan Berant, and Lior Wolf. 2017.
\newblock Language generation with recurrent generative adversarial networks
  without pre-training.
\newblock \emph{CoRR}, abs/1706.01399.

\bibitem[{Radford et~al.(2015)Radford, Metz, and Chintala}]{dcgan}
Alec Radford, Luke Metz, and Soumith Chintala. 2015.
\newblock Unsupervised representation learning with deep convolutional
  generative adversarial networks.
\newblock \emph{CoRR}, abs/1511.06434.

\bibitem[{Rajeswar et~al.(2017)Rajeswar, Subramanian, Dutil, Pal, and
  Courville}]{raj}
Sai Rajeswar, Sandeep Subramanian, Francis Dutil, Christopher~Joseph Pal, and
  Aaron~C. Courville. 2017.
\newblock Adversarial generation of natural language.
\newblock \emph{CoRR}, abs/1705.10929.

\bibitem[{Salimans et~al.(2016)Salimans, Goodfellow, Zaremba, Cheung, Radford,
  and Chen}]{improveTrainGAN}
Tim Salimans, Ian Goodfellow, Wojciech Zaremba, Vicki Cheung, Alec Radford, and
  Xi~Chen. 2016.
\newblock Improved techniques for training gans.
\newblock In \emph{Proceedings of the 30th International Conference on Neural
  Information Processing Systems}, NIPS'16.

\bibitem[{Sharma et~al.(2017)Sharma, El~Asri, Schulz, and Zumer}]{nlgeval}
Shikhar Sharma, Layla El~Asri, Hannes Schulz, and Jeremie Zumer. 2017.
\newblock \href {http://arxiv.org/abs/1706.09799} {Relevance of unsupervised
  metrics in task-oriented dialogue for evaluating natural language
  generation}.
\newblock \emph{CoRR}, abs/1706.09799.

\bibitem[{{Srivastava} et~al.(2017){Srivastava}, {Valkov}, {Russell},
  {Gutmann}, and {Sutton}}]{veeGAN}
A.~{Srivastava}, L.~{Valkov}, C.~{Russell}, M.~U. {Gutmann}, and C.~{Sutton}.
  2017.
\newblock \href {http://arxiv.org/abs/1705.07761} {{VEEGAN: Reducing Mode
  Collapse in GANs using Implicit Variational Learning}}.
\newblock \emph{ArXiv e-prints}.

\bibitem[{Yang et~al.(2018)Yang, Hu, Dyer, Xing, and Berg{-}Kirkpatrick}]{yang}
Zichao Yang, Zhiting Hu, Chris Dyer, Eric~P. Xing, and Taylor
  Berg{-}Kirkpatrick. 2018.
\newblock \href {http://arxiv.org/abs/1805.11749} {Unsupervised text style
  transfer using language models as discriminators}.
\newblock \emph{CoRR}, abs/1805.11749.

\bibitem[{Young et~al.(2017)Young, Hazarika, Poria, and Cambria}]{recent}
Tom Young, Devamanyu Hazarika, Soujanya Poria, and Erik Cambria. 2017.
\newblock Recent trends in deep learning based natural language processing.
\newblock \emph{CoRR}, abs/1708.02709.

\bibitem[{Yu et~al.(2016)Yu, Zhang, Wang, and Yu}]{seqgan}
Lantao Yu, Weinan Zhang, Jun Wang, and Yong Yu. 2016.
\newblock Seqgan: Sequence generative adversarial nets with policy gradient.
\newblock \emph{CoRR}, abs/1609.05473.

\bibitem[{{Zhao} et~al.(2017){Zhao}, {Kim}, {Zhang}, {Rush}, and
  {LeCun}}]{arae}
J.~{Zhao}, Y.~{Kim}, K.~{Zhang}, A.~M. {Rush}, and Y.~{LeCun}. 2017.
\newblock \href {http://arxiv.org/abs/1706.04223} {{Adversarially Regularized
  Autoencoders}}.
\newblock \emph{ArXiv e-prints}.

\bibitem[{Zhu et~al.(2015)Zhu, Kiros, Zemel, Salakhutdinov, Urtasun, Torralba,
  and Fidler}]{bookCorpus}
Yukun Zhu, Ryan Kiros, Richard~S. Zemel, Ruslan Salakhutdinov, Raquel Urtasun,
  Antonio Torralba, and Sanja Fidler. 2015.
\newblock Aligning books and movies: Towards story-like visual explanations by
  watching movies and reading books.

\end{thebibliography}
\bibliographystyle{acl_natbib}

\end{document}